\documentclass[10pt,twocolumn,letterpaper]{article}

\usepackage[pagenumbers]{cvpr} 

\usepackage[dvipsnames]{xcolor}

\definecolor{cvprblue}{rgb}{0.21,0.49,0.74}
\usepackage[pagebackref,breaklinks,colorlinks,citecolor=cvprblue]{hyperref}

\usepackage{amsmath}
\usepackage{amssymb}
\usepackage{booktabs}
\usepackage{multirow}
\usepackage{mathtools, cuted}

\def\confYear{2024\xspace}

\title{DynamicSurf: Dynamic Neural RGB-D Surface Reconstruction with an Optimizable Feature Grid}

\author{
Mirgahney Mohamed \\
Department of Computer Science\\
University College London\\
\and
Lourdes Agapito \\
Department of Computer Science\\
University College London\\
}

\begin{document}
\maketitle
\begin{strip}
\centering
\includegraphics[width = \linewidth]{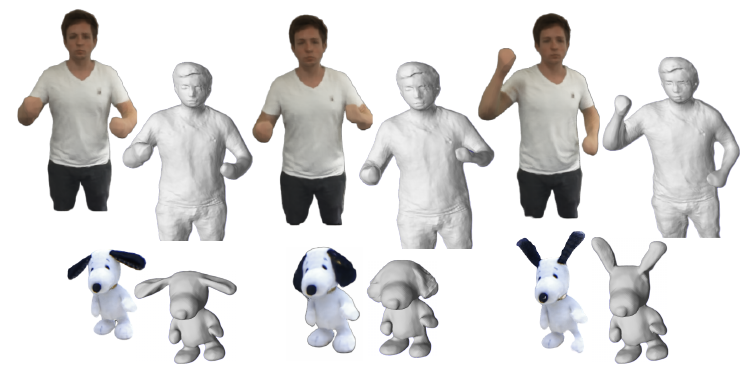}
\captionof{figure}{
\textbf{Reconstruction result on two dynamic sequences:}
            DynamicSurf takes as input a monocular RGB-D sequence of a deforming object and recovers high-fidelity surface reconstructions. }
        \label{fig:teaser}
\end{strip}

\begin{abstract}
We propose DynamicSurf, a model-free neural implicit surface reconstruction method for high-fidelity 3D modelling of non-rigid surfaces from monocular RGB-D video. To cope with the lack of multi-view cues in monocular sequences of deforming surfaces, one of the most challenging settings for 3D reconstruction, DynamicSurf exploits depth, surface normals, and RGB losses to improve reconstruction fidelity and optimisation time. DynamicSurf learns a neural deformation field that maps a canonical representation of the surface geometry to the current frame. We depart from current neural non-rigid surface reconstruction models by designing the canonical representation as a learned feature grid which leads to faster and more accurate surface reconstruction than competing approaches that use a single MLP.  
We demonstrate DynamicSurf on public datasets and show that it can optimize sequences of varying frames with $6\times$ speedup over pure MLP-based approaches while achieving comparable results to the state-of-the-art methods. 
\footnote{Project is available at https://mirgahney.github.io//DynamicSurf.io/.}
\end{abstract}    

\section{Introduction}
\label{sec:intro}

Reconstructing non-rigid surfaces from sequences acquired from a single static viewpoint arguably poses the most challenging scenario for 3D geometry acquisition. 
The small amount of effective multi-view supervision signal present in monocular sequences of deforming scenes, results in ambiguities that are usually tackled by making use of scene or deformation priors. 
While tremendous progress has taken place in the area of model-based approaches to 3D reconstruction of deformable categories such as human bodies, faces or hands (\ie SMPL~\cite{SMPL_2015}, 3DMM~\cite{Blanz1999AMM, zuffi20173d, li2017learning, cao2013facewarehouse}, MANO~\cite{MANO_SIGGRAPHASIA_17}) which fit a pre-trained model to image or depth observations, model-free approaches have received less attention. 

The seminal method  DynamicFusion~\cite{newcombe2015dynamicfusion} was the first to demonstrate model-free real-time reconstruction of 3D surfaces from a live RGB-D sequence. The strategy was to solve for a dense 3D deformation field that maps the current scene geometry to a canonical representation. 
This idea of decomposing a non-rigid scene into a canonical space, and a per frame deformation field was extremely successful and spurred on follow-up work~\cite{innmann2016volumedeform,slavcheva2017killingfusion} with different loss terms or hand-crafted deformation priors. 

Recent years have seen the development of learning-based approaches to 3D reconstruction. Neural radiance fields~\cite{mildenhall2020nerf} encode the density and appearance of scene coordinate points in the weights of a fully connected neural network that is trained in a self-supervised way, purely from images and their associated camera poses. While the focus of neural radiance fields has been their application to novel view synthesis with spectacular success, neural scene representations have also been used for implicit surface reconstruction~\cite{wang2021neus,yariv2020multiview,yariv2021volume,icml2020_2086} by expressing volume density as a function of the underlying 3D surface, leading to improved geometry estimation. Some recent approaches have also exploited the combination of RGB losses with depth cues from RGB-D sequences~\cite{azinovic2022neural,wang2022go-surf} or general-purpose monocular predictors \cite{yu2022monosdf} to successfully resolve reconstruction ambiguities in less-observed or texture-less areas. 

While the original NeRF formulation encodes the scene using a multi-layer perceptron, taking advantage of the smoothness and coherence priors inherently encoded in its architecture, sparse or dense optimizable feature grids have recently become a powerful alternative to overcome the long training times required by MLP-based approaches\cite{yu2021plenoxels,muller2022instant,karnewar2022relu}. 

Neural scene representations have also been extended to model dynamic scenes. The most successful formulations have inherited the canonical space and deformation field decomposition~\cite{pumarola2021d, park2021nerfies, park2021hypernerf} and model scene geometry and appearance by predicting density and colour for each spatial location. NDR~\cite{Cai2022NDR}, like us, relies on RGB-D inputs and applies depth losses to minimize the discrepancies between the rendered and input RGB and depth images. However, all approaches proposed so far for neural non-rigid reconstruction exploit coordinate-based MLPs instead of optimizable feature grids. 

In this paper, we present DynamicSurf a dynamic neural surface reconstruction method from monocular RGB-D input with a static viewpoint. 
To better model the exact location of the deforming surface, we adopt an SDF-based representation and use a differentiable surface rendering pipeline to render both RGB and depth images to supervise surface reconstruction with photometric, depth and surface smoothness losses.
To avoid the slow training times of pure-MLP non-rigid neural RGB-D surface reconstruction~\cite{Cai2022NDR}, 
DynamicSurf utilizes message passing ~\cite{kipf2017semisupervised, veličković2018graph} on dense feature grid ~\cite{wang2022go-surf} which we call \textit{Geometric-feature grid}, and represents the canonical geometry and texture of the deformable objects, using this \textit{Geometric-feature grid}.
We combine the canonical grid representation with a topology-aware network~\cite{park2021hypernerf} to address topological changes.
Our efficient representation and architecture allow for \emph{$(3$-$6)\times$} speed gain to the state-of-the-art methods ~\cite{Cai2022NDR, lin2022occlusionfusion} while maintaining the same level of details and sometimes better.
To the best of our knowledge, DynamicSurf is the first method to bring learnable feature grids to dynamic SDF reconstruction from monocular RGB-D sequences. 
\begin{figure*}[t]
    \includegraphics[width=1\linewidth]{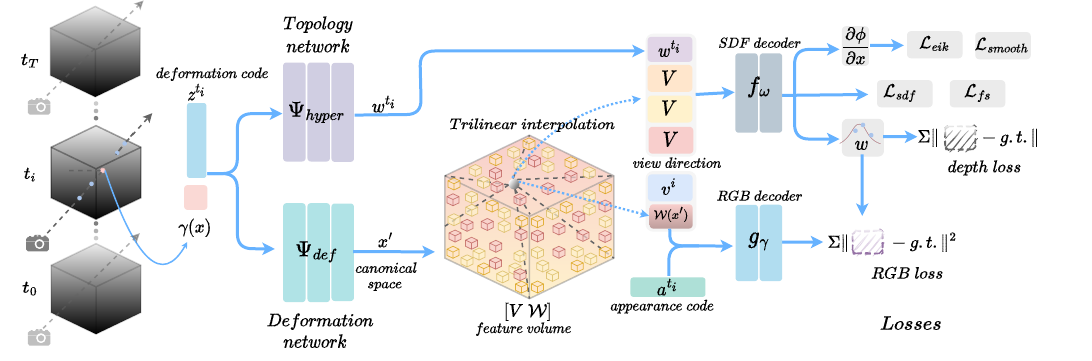}
    \caption{\textbf{Overall architecture of DynamicSurf.} 
    Given a monocular sequence of RGB-D frames and segmentation masks, we learn a deformation network to map points to canonical space and a topology network to model topological changes. 
    Points in canonical space are then queried via trilinear interpolation with our feature grid. 
    These features and the hyper-space features (output of the topology network) are concatenated and decoded with shallow MLPs to predict signed distance values (SDF) of the surface and RGB colors.}
    \label{fig:dynamciSurf}
\end{figure*}

\section{Related Work}
\label{sec:related}
\paragraph{Dynamic RGB reconstruction.}
Template-based methods ~\cite{Blanz1999AMM, SMPL_2015, MANO_SIGGRAPHASIA_17, zuffi20173d, li2017learning, cao2013facewarehouse} are statistical category-specific models learned from  high-quality 3D scans. 
They provide low-dimensional representation that disentangles shape and appearance. 
Utilizing 3D morphable models ~\cite{Blanz1999AMM, li2017learning, cao2013facewarehouse} some methods ~\cite{blanz2003reanimating, cao2014displaced, ichim2015dynamic, guo2018cnn, thies2016face2face} learn to reconstruct heads and faces from RGB information.
While others ~\cite{hong2022headnerf, Grassal_2022_CVPR} in addition to using morphable models they use the recent neural rendering methods ~\cite{mildenhall2020nerf} to model faces from RGB inputs.
Some work ~\cite{bogo2016keep, habermann2019livecap, habermann2020deepcap, jiang2022selfrecon, xu2018monoperfcap, zheng2021pamir} recover digital avatars from monocular 2D input with the help of human parametric models ~\cite{anguelov2005scape,SMPL_2015}.
Despite their success, it still remains challenging to extend model-based methods to general objects with limited 3D scans, such as animals, and humans with diverse clothes and articulated objects.

Non-rigid structure from motion (NR-SFM) algorithms ~\cite{Bregler_2000, Kong_2019_ICCV, sidhu2020neural, Gotardo_2011, Kumar_2020_WACV}  reconstruct class-agnostic objects from 2D trajectories but depend heavily on reliable point trajectories throughout observed sequences ~\cite{Sand_2006, Sundaram_2010}.
Sidhu \etal ~\cite{sidhu2020neural} introduce the first dense neural non-rigid structure from motion (NR-SfM) approach, which is trained end-to-end in an unsupervised manner using an auto-decoder as a deformation model while imposing subspace constraints on the latent space.
LASR ~\cite{yang2021lasr} and  ViSER ~\cite{yang2021viser} recover articulated 3D shapes and dense 3D trajectories from monocular videos using differentiable rendering ~\cite{liu2019softras}.
BANMo ~\cite{Yang_2022_CVPR} merges volumetric neural rendering (NeRFs) with invertible deformation through neural blend skinning, and learns dense correspondences via canonical embeddings.

\paragraph{Dynamic RGB-D reconstruction.}
The seminal method of dynamic RGB-D object reconstruction DynamicFusion~\cite{newcombe2015dynamicfusion} proposes to model dynamic scenes by estimating a dense 6D motion field that warps the state of the scene into a canonical frame. 
VolumeDeform~\cite{innmann2016volumedeform} reduces the inherent drift during motion tracking by using sparse color features SIFT ~\cite{sift_david_2004}. 
KillingFusion~\cite{slavcheva2017killingfusion} is also a templet-free geometry-driven model, estimating a deformation field that aligns signed distance fields (SDFs) representation of the shape. 
While enforcing a local rigidity constraint and approximating a killing vector field.
Guo \etal ~\cite{guo2017real} propose to use shading information to leverage appearance information. 
Then use albedo fusing and geometry to incorporate information for multiple frames.
Similar to KillingFusion ~\cite{slavcheva2017killingfusion} SobolevFusion~\cite{slavcheva2018sobolevfusion} is based on a level set variation method where two SDF fields are aligned with a warping field defined using gradient flow in Sobolev space.

With the success of deep learning, DeepDeform~\cite{bozic2020deepdeform}, based on Siamese networks~\cite{kumar2016learning}, proposes a network that predicts the probability heat map of point correspondences, which are later used to enhance tracking for complex motion capturing.
Bozic \etal ~\cite{bozic2020neural} use convolutional neural networks (CNNs) to predict dense correspondences, which are later used to constrain the optimization using as-rigid-as-possible (ARAP)~\cite{ARAB_2007} priors.
OcclusionFusion~\cite{lin2022occlusionfusion}  proposes an LSTM-involved graph neural network (GNN) to infer the motion of occluded regions by exploiting visibility and temporal information.

\paragraph{Dynamic neural radiance fields.}
Following the success of neural radiance fields NeRF ~\cite{mildenhall2020nerf} for novel view synthesis, recent works have extended it to handle dynamic scenes by learning a warping field to deform observations into a shared canonical space ~\cite{pumarola2021d, park2021nerfies, park2021hypernerf, tretschk2021non}, or by modelling a spatiotemporal 4D radiance field ~\cite{li2020neural, gao2021dynamic, Li_2022_CVPR}.
Nerfies~\cite{park2021nerfies} augment NeRF with a \textbf{SE}(3) deformation field that warps observation to a canonical space.
They propose a coarse-to-fine optimization strategy and inspired by physical simulation they also propose an elastic regularization that further improves optimization robustness.
Exploiting a level-set methodology,  HyperNeRF~\cite{park2021hypernerf} extends Nerfies ~\cite{park2021nerfies} by introducing an ambient dimension to express topological changes.
Instead of deforming each sampled point, Tretschk \etal ~\cite{tretschk2021non} model the deformation as ray blending, and  a rigid network to model the static background. 
Both ray blending and rigid networks are trained without explicit supervision.
TiNeuVox ~\cite{tineuvox} uses a feature grid to encode density and colour, and augments their representation with temporal information encoding additional time features with an MLP network.  Unlike our approach, all the methods above take RGB sequences as input and focus on novel view synthesis, while in DynamicSurf we take RGB-D input and focus on accurate surface reconstruction. 

To the best of our knowledge, NDR~\cite{Cai2022NDR} is the only existing method to tackle the problem of surface reconstruction from a monocular RGB-D camera. However, they propose a fully MLP-based representation, which provides smooth reconstructions at the expense of slow training times. Instead, we propose the use of a  feature volume, optimized in a coarse-to-fine fashion, which results in an increase of $6 \times$ speedup.
\section{Methodology}
\label{sec:method}

Given a sequence of RGB-D images $\{(I^t, D^t), t =1,...,T\}$, captured by a static monocular RGB-D camera and segmentation masks $\mathcal{M}^t$ obtained with an off-the-shelf video segmentation approach~\cite{cheng2021mivos, Lin_2022_WACV}, we formulate dynamic surface reconstruction as an optimisation problem.
Our model learns to map the input sequence into a canonical hyper-space $\mathbb{R}^{3+m}$ composed of 3D canonical and topology networks similar to ~\cite{park2021hypernerf, Cai2022NDR}, where $m$ represents the dimensionality of the topology space.
Scene geometry and colour are represented using feature grids, which are decoded into SDF and RGB values with two shallow MLP decoders shared across grids ~\cite{wang2022go-surf}.
Unlike other approaches, we do not utilize any structural priors such as 2D annotations~\cite{DeepDeform_Bozic}, estimated normal maps ~\cite{SelfRecon_Jiang_CVPR22} or optical flow ~\cite{BANMo_Yang_CVPR22}.
Fig. \ref{fig:dynamciSurf} shows an overview of our architecture.

\subsection{Deformation Field}
For a 3D point sampled from $t$-th frame, we learn a continuous deformation field to map the points to a canonical space with our deformation network $\Psi_{def}$.
We formulate $\Psi_{def}$ as a \textbf{SE}(3) field conditioned on a learnable deformation code $z^t$ ~\cite{park2021nerfies}.
More formally, given an input point $x^t_i \in \mathbb{R}^3$ -- it is important to note $x^t_i$ can represent both surface and free-space points.
We encode the fields with a rotation around an anchor point $a$ using log-quaternion $q$ and a displacement $d$ ~\cite{park2021nerfies}.
We use an MLP architecture to model the deformation field $\Psi_{def}: (x,z^t) \rightarrow (r,a,d)$.
The deformation network predicts rotation vector $r$,  anchor point $a$, and a displacement vector $d$.
We use log-quaternion $(0,r)$ and represent valid rotations using the exponent which is guaranteed to be a unit quaternion:
\begin{equation}
    q = exp\big ( \frac{cos \|r \|}{\frac{r}{\|r\|}sin\|r\|} \big)
\end{equation}
Finally, the deformation is given by the following equation:
\begin{equation}
    x'^t_i = q (x^t_i - a)q^{-1} + a + d
\end{equation}
Dynamic scenes may undergo topological changes which can make it hard for the deformation network alone to learn.
Therefore, we augment our deformation field with a topology-aware network ~\cite{park2021hypernerf}.
Our topology network takes as input 3D point $x^t_i$ and the learnable deformation code $z^t$ and learns to map it into $w^t_i$ in the hyper-space $\mathbb{R}^m$.
The corresponding coordinate of $x^t_i$ in the canonical hyper-space is:
\begin{equation}
    p = [x'^t_i, w^t_i] = [\Psi_{def}(x^t_i, z^t), \Psi_{hyper}(x^t_i, z^t)] \in \mathbb{R}^{3 + m}
\end{equation}
where $m$ is the dimension of the topological hyper-space.

\subsection{Canonical Grid  Representation}
\label{sec:grid_representation}
While most feature-grid-based approaches choose multi-resolution feature volumes~\cite{fang2022fast,wang2022go-surf},   we use a single feature grid to encode the canonical scene geometry and shared shallow MLP decoders, but we employ a coarse-to-fine strategy on the grid resolution.
The canonical scene geometry  a one-level feature grid $\mathcal{V}_{\theta} = \{V^{l}\}$, where $l \in \{0, 1, 2\}$ represents the current level used within the coarse-to-fine strategy.

Given a deformed point $p$ in canonical space $\mathbb{R}^{3 + m}$, we obtain its geometric feature by trilinearly interpolating its 3D location $x'^t_i$ features in our feature grid and concatenating its hyper-space coordinate $w^t_i$ to make what we denote \textit{hyper-features}.
The hyper-features are then decoded into an SDF value $\varphi(\mathbf{x})$ via the geometry MLP $f_{\omega}(\cdot)$:
\begin{align} \label{SDF_MLP}
    \varphi(\mathbf{x}^t_i) &= f_{\omega}([V^l(\mathbf{x}'^t_i), w^t_i])
\end{align}
To decode colour information, following ~\cite{Cai2022NDR} we first map the viewing direction into the canonical space using the Jacobian matrix $J_x(x^t_i) = \frac{\partial x'^t_i}{\partial x^t_i}$ of the 3D canonical point $x'^t_i$ w.r.t observed point $x^t_i$.
Then we encode the color using a separate feature grid $\mathcal{W}_{\beta}$ and decoder $g_{\gamma}(\cdot)$ conditioned on the canonical viewing direction, surface normals $n^t_i$ and a global appearance latent code $a^t$ as in ~\cite{park2021nerfies}:
\begin{equation} \label{color_MLP}
    \mathbf{c}^t_i = g_{\gamma}(\mathcal{W}_{\beta}(\mathbf{x}'^t_i), \mathbf{d}_c, n^t_i, a^t)
\end{equation}
where $\mathbf{d}_c = J_x(x^t_i) \mathbf{d}$ is the viewing direction in  canonical space, $n^t_i = \nabla \varphi(x^t_i)$ is the surface normal at point $x^t_i$.

\begin{figure*}[t]
    \centering
    \includegraphics[width = 1\linewidth]{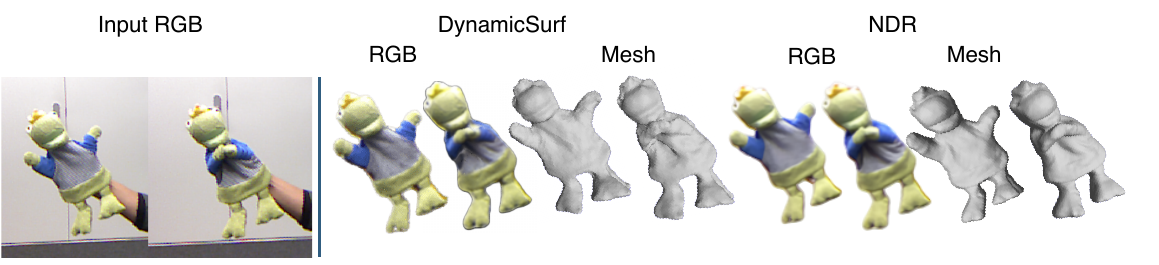}
    \caption{Qualitative comparison with NDR ~\cite{Cai2022NDR}. 
    The result shows our method is on par with NDR ~\cite{Cai2022NDR} while enjoying $6\times$ speed gain.}
    \label{fig:frog}
\end{figure*}

\paragraph{Coarse-to-Fine strategy on the grid resolution.}
Unlike MLP-based methods which enjoy natural smoothness, grid-based methods can suffer from noise and surface artefacts and are prone to fall into suboptimal local minima ~\cite{liu2022devrf}.
To circumvent these problems and to better enhance feature grid smoothness, we use a coarse-to-fine optimization approach ~\cite{liu2022devrf}, progressively increasing the feature grid resolution from $35^3$ to $140^3$.
The lowest resolution models the overall surface, and during training, we increase the grid resolution via interpolation to capture finer surface details
\begin{align} \label{ctf_feature_grid}
    V^{l} &= interp(., V^{l - 1})
\end{align}
where $.$ represents the 3D coordinates of points at the previous resolution level and $V^{l-1}$ and $V^{l}$ denote the features at the previous and current resolution respectively.
\subsection{Depth and Color Rendering}
Inspired by the recent work on learnable feature grid ~\cite{wang2022go-surf} and volume rendering~\cite{mildenhall2020nerf}, we adopt rendering equations to account for dynamic scenes.
For frame $t$, we sample points $\{\mathbf{x}^t_i | \mathbf{x}^t_i = \mathbf{o} + d_i\mathbf{r}, i = 1, \dots, N\}$ along ray $r$ parameterised by camera centre $\mathbf{o}$ and ray direction $d_i$.
We use unbiased and occlusion-aware weights $w^t_i = T^t_i\alpha^t_i$ \cite{wang2021neus}, and $T^t_i = \prod_{j=1}^{i-1}(1-\alpha^t_j)$ represents the \textit{accumulated transmittance} at point $\mathbf{x}^t_i$, and $\alpha^t_i$ is the \textit{opacity value} defined by:
\begin{equation} \label{volume_rendering_alpha}
    \alpha^t_i = \max \bigg(\frac{\mathcal{T}_{\lambda}(\varphi(\mathbf{x}^t_{i})) - \mathcal{T}_{\lambda}(\varphi(\mathbf{x}^t_{i+1}))}{\mathcal{T}_{\lambda}(\varphi(\mathbf{x}^t_{i}))}, 0 \bigg)
\end{equation}
where $\mathcal{T}_{\lambda}(x) = (1 + e^{-\lambda x})^{-1}$ is the Sigmoid function modulated by a learnable parameter $\lambda$ which controls surface transition.
The expected values of predicted colors $\mathbf{c}^t_r$ and sampled depth $d^t_r$ become:
\begin{equation} \label{volume_rendering}
    \hat{\mathbf{c}}^t_r = \sum_{i=1}^{N}w^t_i \hat{\mathbf{c}}^t_{ri}, \quad \hat{d}^t_r = \sum_{i=1}^{N}w^t_i \hat{d}^t_{ri}
\end{equation}
The RGB and depth per-ray rendering losses are:
\begin{equation} \label{eq:rendering}
    \ell^r_{rgb} = \lVert \mathbf{c}^t_r - \hat{\mathbf{c}}^t_r \rVert, \quad \ell^r_{d} = \lvert d^t_r - \hat{d}^t_r \rvert
\end{equation}
where $\mathbf{c}^t_r$ and $d^t_r$ represent the ray $r$ RGB value in the image and the corresponding depth map value respectively.

\begin{figure*}[t]
    \centering
    \includegraphics[width = 1\linewidth]{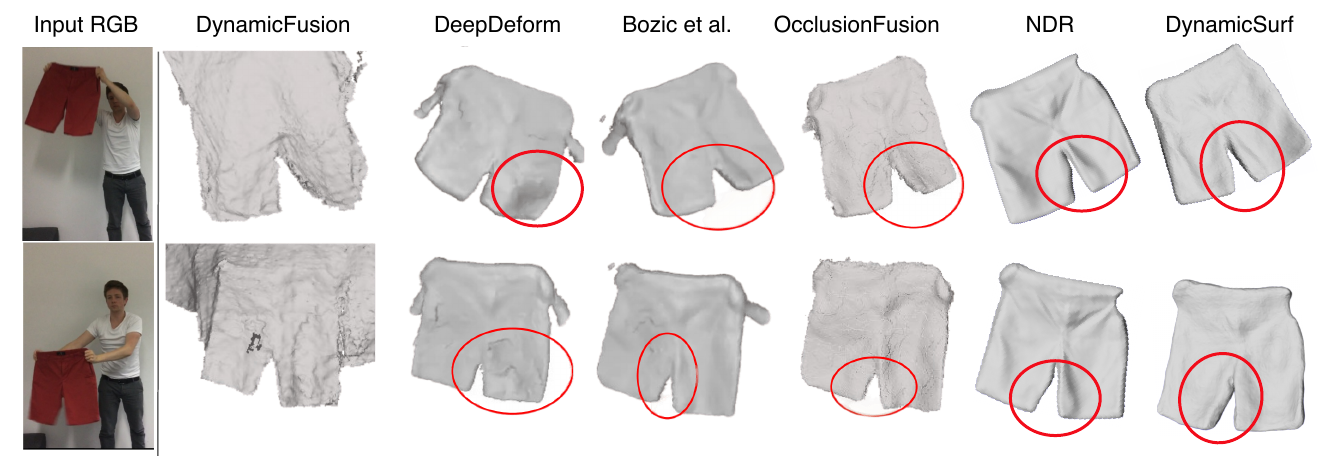}
    \caption{We compare aganist NDR ~\cite{Cai2022NDR}, OcclusionFusion ~\cite{lin2022occlusionfusion}, Bozic \etal ~\cite{bozic2020neural}, DeepDeform ~\cite{bozic2020deepdeform} and DynamicFusion ~\cite{newcombe2015dynamicfusion} on the \textit{shorts} sequence from DeepDeform ~\cite{bozic2020deepdeform} dataset.
    As the code is not available the results of DeepDeform ~\cite{bozic2020deepdeform} and Bozic ~\cite{bozic2020neural} \etal are taken from the video of Bozic \etal ~\cite{bozic2020neural}
    The qualitative comparison shows our method achieves comparable results to NDR ~\cite{Cai2022NDR} and better results than all the other baselines.
    }
    \label{fig:short}
\end{figure*}

\subsection{SDF Supervision and Regularization}
\paragraph{SDF supervision.}
Similar to \cite{wang2022go-surf, azinovic2022neural, Ortiz:etal:iSDF2022}, we approximate the ground truth SDF value based on distance to the observed depth along ray direction $d^t_r$. 
We define the bound $b_r(\mathbf{x^t_i}) = d^t_r - d(x^t_i)$ and divide the set of sampled points into two disjoint sets: near-surface points $S^r_{tr} = \{x^t_i | b_r(x^t_i) <= \epsilon\}$ and free-space points (far from the surface) $S^r_{fs} = \{x^t_i | b_r(x^t_i) > \epsilon\}$.
The truncation threshold $\epsilon$ is a hyper-parameter.
For the set of near-surface points along the ray $S^r_{tr}$ we apply the following SDF loss:
\begin{equation} \label{near_loss}
\begin{split}
     \mathcal{L}^r_{sdf} = \frac{1}{|S^r_{tr}|} \sum_{s \in S_{tr}} |\varphi(\mathbf{x}^t_s) - b_r(\mathbf{x}^t_s)|
\end{split}
\end{equation}
For the set of points far from the surface $S_{fs}$ we apply the free space loss as in ~\cite{wang2022go-surf, Ortiz:etal:iSDF2022} to encourage free space prediction and provide more direct supervision than the rendering terms in \ref{eq:rendering}.
\begin{equation} \label{far_loss}
\begin{split}
    \mathcal{L}^r_{fs} = \frac{1}{|S^r_{fs}|} \sum_{s \in S_{fs}} \max \Big(0, e^{-\alpha \varphi(\mathbf{x}^t_s)} - 1, \varphi(\mathbf{x}^t_s) - b_r(\mathbf{x}^t_s)\Big) 
\end{split}
\end{equation}
An exponential penalty is applied for negative SDF values, a linear penalty for positive SDF values larger than the bound and no penalty is applied if it is smaller.

\paragraph{SDF regularization.}
To encourage valid SDF values, especially in areas without direct supervision we employ the Eikonal regularisation term $\ell_{eik}$, which encourages a uniform increase in the absolute value of SDF as we move far from the surface \cite{icml2020_2086, wang2021neus, Ortiz:etal:iSDF2022}. 
More formally for any query point $x'^t_i$ in the canonical space $\mathbb{R}^3$, we encourage the gradient of the SDF value w.r.t. the 3D observed point to have unit length:
\begin{equation} \label{eq:eikonal_loss}
    \begin{split}
    \mathcal{L}^r_{eik} = \frac{1}{|S^r_{fs}|} \sum_{s \in S_{fs}} \left(1 - \lVert\nabla \varphi(\mathbf{x'^t_s})\rVert\right)^2 
    \end{split}
\end{equation}
\paragraph{Surface smoothness regularization.}
To further enhance surface smoothness we enforce nearby points to have similar normals. Unlike ~\cite{wang2022go-surf}, where they sample uniformly inside the grid, we sample surface points only $x^t_s \in S_{surf}$ which reduces the computation cost drastically.
\begin{equation}
    \begin{split}
    \mathcal{L}_{smooth} = \frac{1}{R} \sum_{s \in S_{surf}} \| \nabla \varphi(x^t_s) - \nabla \varphi(x^t_s + \delta) \|^2 
    \end{split}
\end{equation}
where $x^t_s$ is back-projected using depth maps, $\delta$ is a small perturbation sampled from a uniform Gaussian distribution with standard deviation $\delta_{std}$, and $R$ is the total number of sampled rays.

\subsection{Optimization}
From a randomly sampled frame $t$ we  sample a batch of rays $R$ from all pixels across the image and we minimize the following loss function $\mathcal{L}$:
\begin{equation}
\begin{split}
    \mathcal{L} = \lambda_{rgb}\mathcal{L}_{rgb} + \lambda_{d}\mathcal{L}_{d} + \lambda_{sdf}\mathcal{L}_{sdf} + \lambda_{fs}\mathcal{L}_{fs} + \\ \lambda_{eik}\mathcal{L}_{eik} + \lambda_{smooth}\mathcal{L}_{smooth} + \lambda_{mask} \mathcal{L}_{mask}
\end{split}
\end{equation}
The RGB rendering loss $\mathcal{L}_{rgb}$ measures the difference between the ground truth ray colour and predicted colour over all sampled rays, while the depth rendering loss $\mathcal{L}_{d}$ measures the difference between ground-truth depth and predicted depth values over rays with valid depth $R_d$.
\begin{equation}
    \mathcal{L}_{rgb} = \frac{1}{|R_{rgb}|} \sum_{r \in R_{rgb}} \mathcal{M}^t_r \ell_{rgb}^r, \quad \mathcal{L}_{d} = \frac{1}{|R_d|} \sum_{r\in R_d} \mathcal{M}^t_r \ell_{d}^r
\end{equation}
$\mathcal{M}_r$ is the object mask, which we also utilize as a mask loss to better focus on the object of interest.
\begin{equation}
    \mathcal{L}_{mask} =  H(\mathcal{M}^t_r, \hat{\mathcal{M}}^t_r)
\end{equation}
where $\hat{\mathcal{M}}^t_r = \sum T^t_i\sigma(x^t_i)$ is the density accumulated along ray $r$ and $H$ is the binary cross entropy function.
The SDF loss $\mathcal{L}_{sdf}$ is applied on sampled points inside the truncation region $S_{tr}$.
\begin{equation}
    \mathcal{L}_{sdf} = \frac{1}{|R_d|} \sum_{r=1}^{R} \mathcal{L}^r_{sdf}
\end{equation}
The free-space and Eikonal losses $\mathcal{L}_{fs}$, $\mathcal{L}_{eik}$ are applied to the rest of the points $S_{fs}$:
\begin{eqnarray}
    \mathcal{L}_{fs} = \frac{1}{|R_d|} \sum_{r=1}^{R} \mathcal{L}^r_{fs} \\
    \mathcal{L}_{eik} = \frac{1}{|R_d|} \sum_{r=1}^{R}\mathcal{L}^r_{eik}
\end{eqnarray}

\paragraph{Geometric Initialisation.}
We found it very important to initialize our hyper-feature grid and geometry decoder to predict a sphere~\cite{icml2020_2086, wang2022go-surf} centred at volume origin and with a radius proportional to the scale of the object of interest.
\begin{figure*}[t]
    \centering
    \includegraphics[width = 1\linewidth]{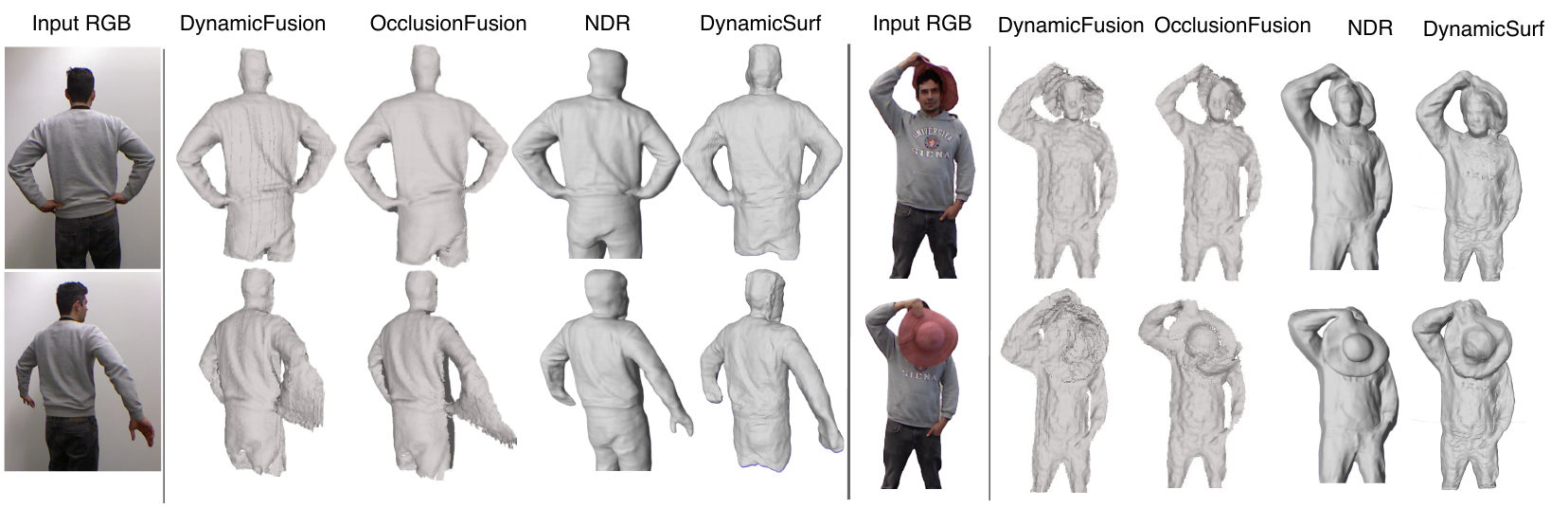}
    \caption{Qualitative comparison with NDR ~\cite{Cai2022NDR}, OcclusionFusion ~\cite{lin2022occlusionfusion}, and DynamicFusion ~\cite{newcombe2015dynamicfusion} on two sequences from the KillingFusion ~\cite{slavcheva2017killingfusion} dataset. 
    The Qualitative comparison shows our method achieves comparable results to the baselines while being $6.7 \times$ faster.
    }
    \label{fig:alex_hat}
\end{figure*}

\begin{figure*}[t]
    
    \centering
    \includegraphics[width = 1\linewidth]{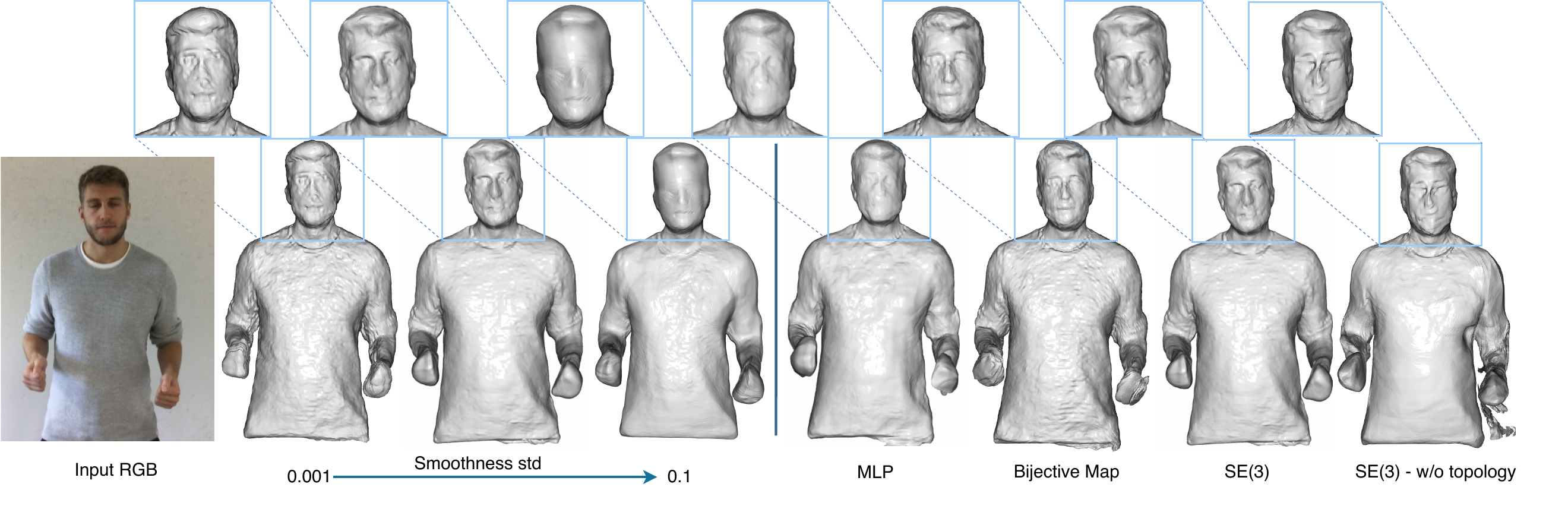}
    \caption{Ablation on smoothness loss and model architecture. Left: shows the effect of standard deviation $\delta_{std}$ on the surface smoothness. 
    Right: shows the importance of deformation and topology networks in recovering surface details and capturing motion.
    }
    \label{fig:ablation}
\end{figure*}
\section{Experimental Evaluation}
\label{sec:experiments}
\paragraph{Implementation details.}
We train DynamicSurf using Adam optimizer ~\cite{kingma2014adam} with a learning rate $5e^{-4}$.
And step based weighting for the RGB, depth and other regularization loss, where we give more weight to the RGB and other regularization losses at the beginning of the training and we reduce that gradually toward the end.
First, we randomly sample an image, then we sample 1024 rays per batch with 128 points along each ray.
Similar to ~\cite{Cai2022NDR, wang2021neus} after we sample uniformly 64 points, we iteratively use importance sampling 4 times for 16 points each for a total of 128 points.
At the beginning of the training, we start with $35^3$ feature grid resolution and $32$-dim features. 
We double the grid size twice during the training using trilinear interpolation as discussed in \cref{sec:grid_representation}.
Out of the 32 feature dimensions we use 6 dimensions for colour decoder and the remaining for the SDF decoder with an ambient dimension of 2 for KIllingFusion ~\cite{slavcheva2017killingfusion} sequences and 8 for DeepDeform ~\cite{bozic2020deepdeform} sequences.
We use a two-layered MLP network for both colour and SDF with $64$ neurons each.
And for the topology-aware network with follow the same architecture as in HyperNeRF ~\cite{park2021hypernerf}.
We also utilize a coarse-to-fine positional encoding strategy, as in Nerfies ~\cite{park2021nerfies}.
Following ~\cite{Cai2022NDR} we leaverage off-the-shelf segmentation methods for humans ~\cite{Lin_2022_WACV} and objects ~\cite{cheng2021mivos} and we apply Robust ICP~\cite{zhang2021fast} for per-frame pose initialization.
All experiments were trained on a single NVIDIA Tesla V100 GPU for a total of 80K for DeepDeform ~\cite{bozic2020deepdeform} sequences and 60K for KIllingFusion ~\cite{slavcheva2017killingfusion} sequences.

\paragraph{Baseline methods.}
We compare with DynamicFusion~\cite{newcombe2015dynamicfusion}, two recent learning-based methods ~\cite{bozic2020deepdeform, bozic2020neural} 
both are fusion-based methods and utilize neural networks to learn correspondences.
and two more state-of-the-art neural reconstruction methods: OcclusionFusion ~\cite{lin2022occlusionfusion} 
which estimates the motion of occluded regions from visible ones using a temporal graph neural network (GNN).
and NDR ~\cite{Cai2022NDR}, which models motion and surface reconstruction using neural implicit functions.

\paragraph {Datasets.}
We evaluate our method and baselines on RGB-D videos from the DeepDeform~\cite{bozic2020deepdeform} and  KillingFusion~\cite{slavcheva2017killingfusion} datasets.
The DeepDeform ~\cite{bozic2020deepdeform} dataset is captured with a Structure Sensor mounted on an iPad.
The RGB-D stream is captured and aligned at $640 \times 480$ and $30$ frames per second
Although the data also provides sparse annotated correspondences and scene flow we did not utilize any of them.
The KillingFusion ~\cite{slavcheva2017killingfusion} dataset is collected using Kinect v1 and aligned at $640 \times 480$  resolution.
We use 6 scenes from DeepDeform ~\cite{bozic2020deepdeform}, and 7 scenes from the KillingFusion~\cite{slavcheva2017killingfusion} dataset, containing human motion, animals, objects, clothes and toys.

\paragraph{Evaluation metrics.}
Following previous work ~\cite{bozic2020deepdeform, bozic2020neural, lin2022occlusionfusion, Cai2022NDR} for quantitative evaluation we computed geometric errors by comparing the reconstructed geometry to the depth values inside the object mask.

\subsection{Reconstruction Quality}
\paragraph{Quantitative evaluation.}
We evaluate DynamicSurf following previous work ~\cite{bozic2020deepdeform, bozic2020neural, lin2022occlusionfusion, Cai2022NDR} using the geometric reconstruction error as a metric and the evaluation protocol described in~\cite{bozic2020deepdeform}. 
~\cref{tab:reco_gerror} shows a comparison with NDR~\cite{Cai2022NDR}, which we trained using their code release and settings, on the same datasets as DynamicSurf.
For completeness, we report the results we obtained using their code (marked with * in ~\cref{tab:reco_gerror}), and the results from their paper.
As shown in~\cref{tab:reco_gerror}, DynamicsSurf achieves reconstruction errors on par with NDR ~\cite{Cai2022NDR} 
while being $5-6.7\times$ faster as shown in~\cref{tab:optim_time}. 

\begin{table}[!h]
\centering
\resizebox{0.5\textwidth}{!}{
\begin{tabular}{c|l|clcccc}
\hline
\multirow{3}{*}{Metric}                      & \multicolumn{1}{c|}{\multirow{3}{*}{Method}} & \multicolumn{4}{c}{Dataset}              \\ \cline{3-6} 
 &
  \multicolumn{1}{c|}{} &
  \multicolumn{1}{c|}{DeepDeform} &
  \multicolumn{3}{c}{KillingFusion}
  \\ \cline{3-6} 
 &
  \multicolumn{1}{c|}{} &
  \multicolumn{1}{l|}{Human} &
  \multicolumn{1}{l|}{Alex} &
  \multicolumn{1}{l|}{Hat} &
  \multicolumn{1}{l}{Frog}
  \\ \hline
\multirow{4}{*}{Mean}                        & OcclusionFusion                              & - & 3.75 & 6.77 & 1.61 \\
                                             & NDR                                          & - & 4.24 & 4.93 & 1.34 \\
                                             & NDR$^*$                                        & 3.89 & 5.69 & 3.24 & 2.26 \\
                                             & Ours                                         & 2.96 & 6.12 & 2.83 &  1.39 \\ \hline
\multicolumn{1}{l|}{\multirow{4}{*}{Median}} & OcclusionFusion                              & - & 3.41 & 6.47 & 1.59 \\
\multicolumn{1}{l|}{}                        & NDR                                          & - & 4.11 & 4.66 & 1.33 \\
\multicolumn{1}{l|}{}                        & NDR$^*$                                        & 3.8 & 4.69 & 3.26 & 2.25 \\
\multicolumn{1}{l|}{}                        & Ours                                         & 2.89 & 5.21 & 2.75 & 1.27 \\ \hline
\end{tabular}
}
\caption{Quantitative results on 5 sequences. The geometry error ($\downarrow$) represents the average per-point per-frame 3D error between the reconstructed shape and the point cloud obtained from backprojecting the GT depth values inside the mask. NDR denotes the results from their paper, while NDR$^*$ denotes the result obtained by training on NDR's ~\cite{Cai2022NDR} public code release and settings.}
\label{tab:reco_gerror}
\end{table}

\begin{table}[t]
 \centering
 \resizebox{0.5\textwidth}{!}{
 \begin{tabular}{l|cccccc}
 \hline
 \multicolumn{1}{c|}{\multirow{3}{*}{Method}} & \multicolumn{5}{c}{Dataset} \\ \cline{2-6} 
 \multicolumn{1}{c|}{} &
   \multicolumn{3}{c|}{KillingFusion} &
   \multicolumn{2}{c}{DeepDeform} \\ \cline{2-6} 
   \multicolumn{1}{c|}{} &
   \multicolumn{1}{l|}{Alex} &
   \multicolumn{1}{l|}{Hat} &
   \multicolumn{1}{l|}{Frog} &
   \multicolumn{1}{l|}{Human} &
   \multicolumn{1}{l}{Dog} \\ \hline
 NDR          & 1 (36)  & 1 (37.8) & 1 (36) & 1 (41) & 1 (41.5) \\
 Ours         & $5.8\times$ & $6.7\times$ & $6.2\times$ & $5.1\times$ & $5\times$ \\ \hline
 \end{tabular}
 }
 \caption{Optimization time results on 5 sequences. DynamicSurf archives $(5$-$7)\times$ speed to NDR. 
 The numbers between parentheses show time in hours.
 }
 \label{tab:optim_time}
\end{table}

\paragraph{Qualitative results.}
We present qualitative results of reconstructed sequences in~\ref{fig:frog}, ~\ref{fig:short} and~\ref{fig:alex_hat} which show that DynamicSurf can reconstruct smoother and more complete surfaces than fusion-based methods~\cite{newcombe2015dynamicfusion,lin2022occlusionfusion} and learning based methods~\cite{bozic2020deepdeform,bozic2020neural}  and achieves comparable reconstruction quality to the SOTA method NDR~\cite{Cai2022NDR} while being significantly faster~\cref{tab:optim_time}.

\subsection{Ablation studies} 
To validate our architecture choices and losses, we evaluate three key components of our method based on their contribution to surface reconstruction quality.
Please refer to supplementary materials for more extended evaluations.

\subsubsection{Deformation networks}
We evaluate the reconstruction result with different deformation network architectures.
We replace it with the Bijective map proposed by NDR ~\cite{Cai2022NDR}, and with a pure MLP-based network to model the deformation. 
To keep the comparison fair, we make sure that all networks have the same number of layers and neurons.~\cref{fig:ablation} shows that using our \textbf{SE}(3) deformation network yields better results than the pure MLP-based method, especially in capturing face details.
As for the bijective map~\cite{Cai2022NDR}, while~\cref{fig:ablation} shows comparable reconstruction quality, the bijective map deformation model suffers from surface artefacts, furthermore, it requires high computation cost.
Therefore, we opted for the \textbf{SE}(3) due to its lower computational complexity and for providing smoother surfaces. 

\subsubsection{Topology network}
We train DynamicSurf without the topology network, such that the decoders only take input features from the feature grid. For a fair comparison, we increase the feature volume dimensions to match those of the topology network.
As shown in~\cref{fig:ablation} the network with the topology can better capture the details of the surface, especially with difficult deformations.

\subsubsection{Smoothness loss}
~\cref{fig:ablation} shows the result of varying the standard deviation of the smoothness loss $\delta_{std}$.
As expected, as the value increases, the fine details on the surface start to disappear.
While having a smaller standard deviation results in noise surface see the chest and face of the human in ~\cref{fig:ablation} left.
\section{Limitations} DynamicSurf is sequence specific, since it is an optimization approach, and needs to be trained per  sequence, which can be impractical. 
Further, it does not utilize the shared information between identities across sequences and similar motions.
Also similar to the baselines, DynamicsSurf requires per-frame pose initialization for sequences with global rotation. Learning the global rotations would bring the method a step closer to online processing.
We will investigate how to overcome these limitations in the future. In terms of broader impact, training these models still takes substantial computational and energy resources. Pushing further in the direction of designing light-weight models is an important research direction. 

\section{Conclusions}
\label{sec:conclusions}
We have proposed DynamicSurf a template-free method for high-fidelity surface and motion reconstruction of deformable scenes from monocular RGB-D sequences.
DynamicSurf maps a canonical representation of the surface geometry to the current frame using a neural deformation field.
We represent the canonical space as a learned feature grid, optimized in a coarse-to-fine fashion, and we employ a topology-aware network to handle topology variations.

DynamicSurf achieves comparable reconstruction performance to the state-of-the-art -- purely MLP-based -- non-rigid surface reconstruction method~\cite{Cai2022NDR} on different object categories from various public datasets, while being an order of magnitude faster.

\vspace{-0.1cm}
\section*{Acknowledgements}
\vspace{-0.1cm}
Research presented here has been supported by the UCL Centre for Doctoral Training in Foundational AI under UKRI grant number EP/S021566/1.

{
    \small
    \bibliographystyle{ieeenat_fullname}
    \bibliography{main}
}

\end{document}